\documentclass[final]{cvpr}

\usepackage{times}
\usepackage{epsfig}
\usepackage{graphicx} 
\usepackage{amsmath}
\usepackage{amssymb}
\usepackage{multirow}
\usepackage{makecell}
\usepackage{array}
\usepackage{cite}

\usepackage[pagebackref=true,breaklinks=true,colorlinks,bookmarks=false]{hyperref}

\def\cvprPaperID{7880} 
\def\confYear{CVPR 2022}

\begin{document}

\title{Multi-modal Sensor Fusion for Auto Driving Perception: A Survey}

\author{Keli Huang\thanks{~This work was done during the author's intership in Shanghai AI Lab}\\
University of California\\
Los Angeles, US\\
{\tt\small kelihuang@cs.ucla.edu}
\and
Botian Shi\\
Shanghai AI Lab\\
Shanghai, China\\
{\tt\small shibotian@pjlab.org.cn}
\and
Xiang Li\\
Beijing Institute of Technology\\
Beijing, China\\
{\tt\small 3120211007@bit.edu.cn}
\and
Xin Li\\
East China Normal University\\
Shanghai, China\\
{\tt\small 51194506020@stu.ecnu.edu.cn}
\and
Siyuan Huang\\
Shanghai AI Lab\\
Shanghai, China\\
{\tt\small huangsiyuan@pjlab.org.cn}
\and
Yikang Li\\
Shanghai AI Lab\\
Shanghai, China\\
{\tt\small liyikang@pjlab.org.cn}
}

\maketitle

\begin{abstract}

Multi-modal fusion is a fundamental task for the perception of an autonomous driving system, which has recently intrigued many researchers. However, achieving a rather good performance is not an easy task due to the noisy raw data, underutilized information, and the misalignment of multi-modal sensors. In this paper, we provide a literature review of the existing multi-modal-based methods for perception tasks in autonomous driving. Generally, we make a detailed analysis including over 50 papers leveraging perception sensors including LiDAR and camera trying to solve object detection and semantic segmentation tasks. Different from traditional fusion methodology for categorizing fusion models, we propose an innovative way that divides them into two major classes, four minor classes by a more reasonable taxonomy in the view of the fusion stage. Moreover, we dive deep into the current fusion methods, focusing on the remaining problems and open-up discussions on the potential research opportunities. In conclusion, what we expect to do in this paper is to present a new taxonomy of multi-modal fusion methods for the autonomous driving perception tasks and provoke thoughts of the fusion-based techniques in the future.
 
\end{abstract}

\section{Introduction}

Perception is an essential module for autonomous cars \cite{Geiger2012CVPR, sun2020scalability, li2020deep }. The tasks include but are not limited to 2D/3D object detection, semantic segmentation, depth completion, and prediction, which rely on sensors installed on the vehicles to sample the raw data from the environment. Most existing methods \cite{li2020deep}, conduct the perception tasks on point cloud and image data captured by LiDAR and camera separately, showing some promising achievements.

However, perception by single-modal data suffers from inherent drawbacks \cite{bijelic2020seeing, Geiger2012CVPR}. For example, camera data is mainly captured in the lower position of the front view \cite{yoo20203d}. Objects may be occluded in more complex scenes, bringing severe challenges to object detection and semantic segmentation. Moreover, limited to the mechanical structure, LiDAR has various resolutions at different distances \cite{xie2021x}, and it is vulnerable to extreme weather such as fogs and heavy rains \cite{bijelic2020seeing}. Although the data of the two modalities excel in various areas when used separately \cite{li2020deep}, the complementarity of LiDAR and camera makes the combination result in a better performance on perception\cite{bijelic2020seeing,vora2020pointpainting, xie2020pi}.

Recently, multi-modal fusion methods for perception tasks in autonomous driving have rapidly progressed \cite{wang2021multi, wang2021pointaugmenting, cui2021deep}, varying from more advanced cross-modal feature representations and more reliable sensors in different modalities to more complex and robust deep learning models and techniques of the multi-modal fusion. However, only a few literature reviews \cite{wang2021multi, cui2021deep} concentrate on the methodology of multi-modal fusion methodology itself, and most of them follow a traditional rule of separating them into three major classes as early-fusion, deep-fusion, and late-fusion, focusing on the stage of fusion feature in the deep learning model, whether it is at data-level, feature-level or proposal-level. Firstly, such taxonomy does not make a clear definition of the feature representation in each level. Secondly, it suggests that two branches, LiDAR and camera, are always symmetric in the processing procedure, obscuring the situation fusing proposal-level feature in the LiDAR branch and data-level feature in the camera branch \cite{zhao20193d}. In conclusion, traditional taxonomy may be intuitive but primitive to summarize more and more emerging multi-modal fusion methods recently, which prevents researchers study and analyzing them from a systematic view.

In this paper, we will give a brief review of the recent papers about the multi-modal sensor fusion for autonomous driving perception. We propose an innovative way to divide over 50 related papers into two major classes and four minor classes by a more reasonable taxonomy from the fusion stage perspective.

The main contribution of this work can be summarized as the following:

\begin{itemize}

\item We propose an innovative taxonomy of multi-modal fusion methods for autonomous driving perception tasks, including two major classes, as strong-fusion and weak-fusion, and four minor classes in strong-fusion, as early-fusion, deep-fusion, late-fusion, asymmetry-fusion, which is clearly defined by the feature representations of LiDAR branch and camera branch.

\item We conduct an in-depth survey about the data format and representation of the LiDAR and camera branch and discuss their different characteristics.

\item We make a detailed analysis about the remaining problems and introduce several potential research directions about the multi-modal sensor fusion method, which may enlighten future research works.

\end{itemize}

The paper is organized by the following: In section 2, we give a brief introduction to the perception tasks in autonomous driving, including but not limited to object detection, semantic segmentation, as well as several widely-used open dataset and benchmarks. In section 3, we summarize all the data formats as the input of downstream models. Unlike the image branch, the LiDAR branch many vary in the format as the input, including different manually designed features and representations. We then describe in detail in section 4 the fusion methodology, which is an innovative and clear taxonomy of dividing all the current work into two major classes and four minor classes compared to traditional methods. In section 5, we deeply analyze some remaining problems, research opportunities, and plausible future works about the multi-modal sensor fusion for autonomous driving, which we can easily perceive some insightful attempts but still left to be solved. In section 6, we finally conclude the content of this paper.

\begin{figure*}[ht] 
  \centering
  \includegraphics[width=0.7\textwidth]{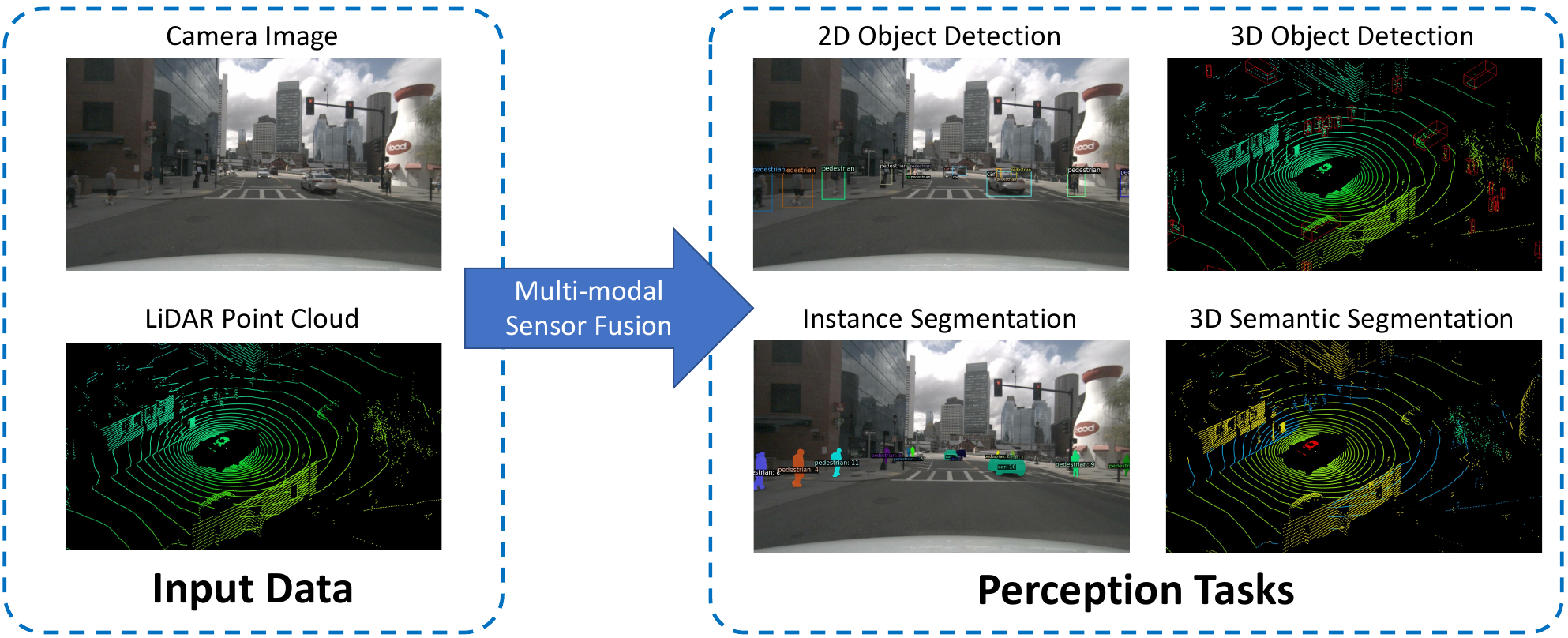}
  \caption{Perception Tasks of Autonomous Driving by Multi-modal Sensor Fusion Model.}
  \label{fig:perception_tasks}
\end{figure*}

\section{Tasks, and Open Competitions}

We will firstly introduce the commonly-known perception tasks in autonomous driving in this section. Besides, there is also some widely-used open benchmark dataset that we will give a glimpse of them here.

\subsection{Multi-modal Sensor Fusion Perception Tasks}

In general, a few tasks can be accounted for as Driving Perception Tasks, containing Object Detection, Semantic segmentation, Depth Completion \& Prediction, and so on\cite{Geiger2012CVPR, sun2020scalability}. Here, we mainly focus on the first two tasks as one of the most concentrated research areas. Also, they cover tasks such as the detection of obstacles, traffic lights, traffic signs, and the segmentation of lanes or free space. We also briefly introduce some remaining tasks. An overview of perception tasks in the autonoumous driving is shown in Figure \ref{fig:perception_tasks}. 

\textbf{Object Detection}

It is crucial for autonomous driving cars to understand the surrounding environment. Unmanned vehicles need to detect both stationary and moving obstacles on the road for safe driving. Object Detection, a traditional computer vision task, is widely-used in autonomous driving systems\cite{zhou2018voxelnet, qi2017pointnet}. Researchers build such framework for Obstacle Detection (Cars, Pedestrians, Cyclist, etc.), Traffic Light Detection, Traffic Sign Detection, and so on. 

Generally speaking, object detection uses rectangles or cuboids represented by parameters to tightly bound instances of predefined categories, as cars or pedestrians, which needs to both excel in localization and classification. Because of the lack of depth channel, 2D object detection is often expressed as $(x,y,h,w,c)$ simply while 3D object detection bounding box is often represented as $(x,y,z,h,w,l,\theta,c)$.

\textbf{Semantic Segmentation}

Except for object detection, many autonomous driving perception tasks can be formulated as semantic segmentation. For example, free-space detection \cite{kemsaram2019integrated, pazhayampallil2014free, zhao2018overview} is a basic module for many autonomous driving systems which classify the ground pixel into the drivable and non-drivable parts. Some Lane Detection \cite{fernandez2014road, wang2018lanenet} methods also use the multi-class semantic segmentation mask to represent the different lanes on the road.

The essence of semantic segmentation is to cluster the basic components of input data, such as pixels and 3D points, into multiple regions containing specific semantic information. Specifically, semantic segmentation is that given a set of data, such as image pixels $D_I = \{d_1,d_2,...,d_n\}$ or LiDAR 3D point clouds $D_L = \{d_1,d_2,...,d_n\}$, and a predefined set of candidate labels $Y = \{y_1,y_2,y_3,...,y_k\}$, we use a model to assign each pixel or point $d_i$ with selected one of the k semantic labels or the probabilities for all.

\textbf{Other Perception Tasks}

Besides the object detection and semantic segmentation mentioned above, the perception task in autonomous driving includes object classification \cite{li2020deep}, depth completion and prediction \cite{Geiger2012CVPR}. Object classification mainly solves the problem of determining the category given point clouds and images through a model. The depth completion and prediction tasks focus on predicting the distance for every pixel in the image from the viewer given LiDAR point cloud and image data. Although these tasks may benefit from multi-modal information, the fusion module is not widely discussed in these fields. As a result, we choose to omit such two tasks in this paper.

Although many other perception tasks are not covered in this paper, most can be regarded as object detection or semantic segmentation variants. Therefore, we focus on these two research works in this paper.

\subsection{Open competitions and Datasets}

Over ten datasets \cite{ma2019trafficpredict, sun2020scalability,hu2021towards,pham20203d,wen2020urbanloco,ligocki2020brno,martinez2020pit30m,wang2020tartanair,chen2020cuhk,yan2019eu,yang2019drivingstereo,chang2019argoverse,gruber2019gated2depth,yogamani2019woodscape,xue2019blvd,patil2019h3d,ramanishka2018toward} are related to autonomous driving percception. However, only three datasets are commonly used, including KITTI\cite{Geiger2012CVPR}, Waymo\cite{sun2020scalability}, and nuScenes\cite{caesar2020nuscenes}. Here we summarize the detailed characteristics of these datasets in Table \ref{tab:datasets}.

\begin{table*}[!htb]
    \begin{center}
        \caption{Survey of Commonly Used Open Dataset and Benchmarks}
        \label{tab:datasets}
        \scalebox{0.7}{
        \begin{tabular}
        {m{2cm}<{\centering}|m{0.7cm}<{\centering} | m{0.9cm}<{\centering} | m{2.5cm}<{\centering}  |m{4cm}<{\centering}|m{2.5cm}<{\centering} |
        m{1cm}<{\centering}|m{1cm}<{\centering} |
        m{2cm}<{\centering} | m{2cm}<{\centering}}

            \hline\hline
            Dataset  & Year  & Hours & LiDARs & Cameras &  Annotated LiDAR Frames & 3D Boxes & 2D Boxes & Traffic Scenario & Diversity  \\
            \hline
            
            KITTI \cite{Geiger2012CVPR} & 2012 & 1.5 & 1 Velodyne HDL-64E & 2 color, 2 grayscale cameras & 15k & 80k & 80k & Urban, Suburban, Highway  & -\\

            \hline
            
            Waymo \cite{sun2020scalability} & 2019 & 6.4 & 5 LiDARs & 5 high-resolution pinhole cameras & 230k & 12M & 9.9M & Urban, Suburban   & Locations \\

            \hline
            
            nuScenes \cite{caesar2020nuscenes} & 2019 & 5.5  & 1 Spinning 32-beams LiDAR & 6 RGB cameras & 40k & 1.4M & - & Urban, Suburban  & Locations, Weather \\
            \hline
            
            ApolloScape	\cite{Huang_2018_CVPR_Workshops} &2018&	2&	2 VUX-1HA laser scanners&	2 front cameras	&144k&	70k	&-	& Urban, Suburban, Highway  &	Weather, Locations\\
            \hline
            
            PandaSet \cite{9565009}&	2021&	-&	1 Mechanical spinning LiDAR and 1 Forward-facing LiDAR& 	5 wide-angle cameras and 1 forward-facing long-focus camera&	6k&	1M&	-	& Urban&	Locations\\
            \hline
            
            EU Long-term \cite{yan2020eu}&	2020&	1 &	2 Velodyne HDL-32E LiDARs and 1  ibeo LUX 4L LiDARs&	2 stereo cameras and 2 Pixelink PL-B742F industrial cameras&	-&	-&	-&	 Urban, Suburban&	Season\\
            \hline
            
            Brno Urban \cite{ligocki2020brno}&	2020&	10 &	2 Velodyne HDL-32e LiDARs&	4 RGB cameras &	-&	-&	-	& Urban, Highway & 	Weather\\
            \hline
            
            A*3D \cite{9197385}&	2020&	55&	1 Velodyne HDL-64ES3 3D-LiDAR&	2 color cameras&	39k&	230k&	-&	 Urban&	Weather \\
            \hline
            
            RELLIS \cite{9561251}&	2021&	-&	1 Ouster OS1 LiDAR and 1 Velodyne Ultra Puck&	1 3D stereo camera and 1 RGB camera&	13k&	-&	-&	Suburban &-\\
            \hline
            
            Cirrus \cite{wang2021cirrus}&	2021&	-&	2 Luminar Model H2 LiDARs&	1 RGB camera&	6k&	100k&	-	 &Urban&-	\\
            \hline
            
            HUAWEI ONCE \cite{mao2021one}&	2021&	144	&1 40-beam LiDAR &	8 high-resolution cameras&	16k	&417k&	769k&	 Urban, Suburban&	Weather, Locations \\

            \hline\hline
        \end{tabular}
        }
    \end{center}
\end{table*}

KITTI \cite{Geiger2012CVPR} open benchmark dataset, as one of the most used dataset for object detection in autonomous driving, contains 2D, 3D, and bird’s eye view detection tasks. Equipped with four high-resolution video cameras, a Velodyne laser scanner, and a state-of-the-art localization system, KITTI collected 7481 training images and 7518 test images as well as the corresponding point clouds. Only three objects are labeled as cars, pedestrians, and cyclists with more than 200k 3D object annotations divided into three categories: easy, medium, and hard by detection difficulty. For the KITTI object detection task, Average Precision is frequently used for comparison. Besides, Average Orientation Similarity is also used to assess the performance of jointly detecting objects and estimating their 3D orientation.

As one of the largest open datasets commonly used for autonomous driving benchmarks, the Waymo \cite{sun2020scalability} open dataset was collected by five LiDAR sensors and five high-resolution pinhole cameras. Specifically, there are 798 scenes for training, 202 for validation, and 150 scenes for the test. Each scene spans 20 seconds with annotations in vehicles, cyclists, and pedestrians. For evaluating 3D object detection tasks, Waymo consists of four metrics: AP/L1, APH/L1, AP/L2, APH/L2. More specifically, AP and APH represent two different performance measurements, while L1 and L2 contain objects with different detection difficulties. As for APH, it is computed similar to AP but weighted by heading accuracy.

NuScenes \cite{caesar2020nuscenes} open dataset contains 1000 driving scenes with 700 for training, 150 for validation, and 150 for the test. Equipped with cameras, LiDAR, and radar sensors, nuScenes annotates 23 object classes in every key-frame, including different types of vehicles, pedestrians, and others. NuScenes uses AP, TP for detection performance evaluation. Besides, it proposed an innovative scalar score as the nuScenes detection score (NDS) calculated by AP, TP isolating different error types.

\section{Representations for LiDAR and Image}

The deep learning model is restricted to the representation of the input. In order to implement the model, we need to pre-process the raw data by an elaborate feature extractor before feeding the data into the model. Therefore, we firstly introduce the representations of both LiDAR and image data, and we will discuss the fusion methodology and models in the later section.

As for the image branch, most existing methods maintain the same format as the raw data for the input of the downstream modules \cite{wang2021multi}. However, the LiDAR branch is highly dependent on data format\cite{li2020deep}, which emphasizes different characteristics and influences downstream model design massively. As a result, we summarize them as point-based, voxel-based, and 2D-mapping-based point cloud data formats suiting heterogeneous deep learning models.

\subsection{Image Representation}

As the most commonly used sensor for data acquisition in 2D or 3D object detection and semantic segmentation tasks, the monocular camera provides RGB images rich in texture information \cite{ataer2013tracking, kim2021spatiotemporal, wang2018fusing}. Specifically, for every image pixel as $(u,v)$, it has a multiple channel feature vector as $ F_{(u,v)} = \{R,G,B, ...\} $ which usually contains the camera capture color decomposing in the red, blue, green channel or other manually designed feature as the gray-scale channel.

However, direct detecting objects in 3D space is relatively challenging because of the limited depth information, which is hard to be extracted by a monocular camera. Therefore, many works\cite{li2019stereo,you2019pseudo,chen20173d} use binocular or stereo camera systems through the spatial and temporal space to exploit additional information for 3D object detection, such as depth estimation, optical flow, and so on. For the extreme driving environment such as night or fog, some work also uses gated or infra-red cameras to improve the robustness\cite{bijelic2020seeing}.

\subsection{Point-based Point Cloud Representation}

As for 3D perception sensors, LiDARs use a laser system to scan the environment and generate a point cloud. It samples the points in the world coordinate system that denotes the intersection of the laser ray and the opacity surface. Generally speaking, the raw data of the most LiDARs is quaternion like $(x,y,z,r)$, which $r$ stands for the reflectance of each point. The different texture leads to different reflectance, which offers additional information in several tasks \cite{huang2017traffic}.

In order to incorporate the LiDAR data, some methods use points directly by a point-based feature extraction backbone\cite{qi2017pointnet, qi2017pointnet++}. However, the quaternion representation of points suffers redundancy or speed drawbacks. Therefore, many researchers \cite{zhou2018voxelnet, lang2019pointpillars, shi2020pv, deng2020voxel} try to transform the point cloud into voxel or 2D projection before feeding it to the downstream modules.

\subsection{Voxel-based Point Cloud Representation}

Some work utilizes 3D CNN by discretizing the 3D space into 3D voxels, representing as $X_v = \{x_1, x_2, x_3 ... x_n\}$, where each $x_i$ stands for a feature vector as $x_i = \{s_i, v_i\}$. $s_i$ stands for the centorid of the voxelized cuboid while $v_i$ stands for some statistical-based local information. 

The local density a commonly used feature defined by the quantity of 3D points in the local voxel \cite{che2017fast,vo2015octree}. Local offset is commonly defined as the offset between the point real-word coordinates and the local voxel centroid. Others may contain local linearity and local curvature\cite{rusu20113d,thomas2018semantic}.

Recent work may consider a more reasonable discretizing way as cylinder-based voxelization\cite{xie2021x}, but Voxel-based Point Cloud Representation, unlike mentioned Point-based Point Cloud Representation above, dramatically reduces the redundancy of the unstructured point cloud \cite{lang2019pointpillars}. Besides,  being able to utilize 3D sparse convolutional techniques, the perception tasks achieve not only faster-training speed but also higher accuracy \cite{lang2019pointpillars,deng2020voxel}.

\subsection{2D-mapping-based Point Cloud Representation}

Instead of proposing a new network structure, some works utilize sophisticated 2D CNN backbones to encode the point cloud. Specifically, they tried to project the LiDAR data into image space as two common types, including camera plane map (CPM) and bird's eye view (BEV) \cite{yang2018pixor,lang2019pointpillars}.

A CPM can be obtained with the extrinsic calibration by projecting every 3D point as $(x,y,z)$ into the camera coordinate system as $(u,v)$. Since the CPM has the same format as the camera image, they can be naturally fused by using the CPM as an additional channel. However, due to the lower resolution of LiDAR after the projection, the feature of many pixels in CPM is corrupted. Hence, some methods have been proposed to up-sample the feature map while others left them blank \cite{ku2018joint, lu2019scanet}.

Unlike CPM that directly projects LiDAR information into front-view image space, the BEV mapping provides an elevated view of a scene from above. It is utilized by the detection and localization tasks for two reasons. Firstly, unlike the camera installed behind the windscreen, most LiDARs are on the vehicle's top with fewer occlusions \cite{Geiger2012CVPR}. Secondly, all objects are placed on the ground plane in the BEV, and models can generate predictions without distortion in length and width \cite{Geiger2012CVPR}. BEV components may vary. Some are directly converted from height, density, or intensity as point-based or voxel-based features \cite{chen2017multi}, while others learn features of the LiDAR information in the pillars through feature extractor modules \cite{lang2019pointpillars}.

\section{Fusion Methodology}

In this section, we will review different fusion methodologies on LiDAR-camera data. In the perspective of traditional taxonomy, all multi-modal data fusion methods can be conveniently categorized into three paradigms, including data-level-fusion (early-fusion), feature-level-fusion (deep-fusion), and object-level-fusion (late-fusion )\cite{wang2021multi, cui2021deep, feng2020deep}. 

The data-level-fusion or early-fusion methods directly fuse the raw sensor data in different modalities by spatial alignment. The feature-level-fusion or deep-fusion methods mix cross-modal data in feature space by concatenation or element-wise multiplication. The object-level-fusion methods combine prediction results of models in each modality and make the final decision.

\begin{figure}[htbp]
\centering
\includegraphics[width=0.48\textwidth]{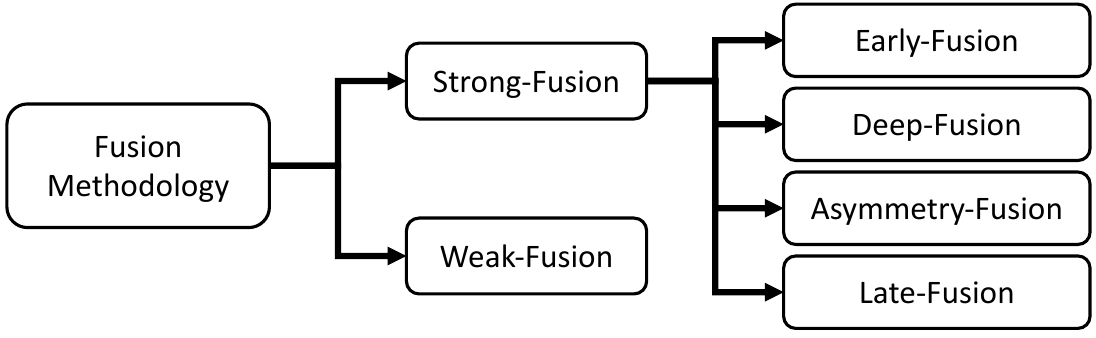}
\caption{Fusion Methodology Overview}
\label{fig:fusion_taxonomy}
\end{figure}

However, recent works \cite{zhao20193d,ku2019improving, deng2019mlod,wang2019frustum,zhang2020faraway} cannot be easily classified into these three categories. So in this paper, we propose a new taxonomy that divides all fusion methods into strong-fusion and weak-fusion, which we will elaborate on in detail. We show their relationships in Figure \ref{fig:fusion_taxonomy}. 

For the performance comparison, we are focusing on two main tasks in KITTI benchmark, as 3D Detection and Bird's Eye View Object Detection. Table \ref{tab:result-bev-kitti} and Table \ref{tab:result-3d-kitti} present the experimental results on KITTI test dataset of BEV and 3D setup separately of recent multi-modal fusion methods.

\begin{table*}[h]
    \begin{center}
        \caption{Survey of BEV Task Results in KITTI\cite{Geiger2012CVPR} for Test Dataset}
        \label{tab:result-bev-kitti}
        \scalebox{0.75}{
        \begin{tabular}
        {m{4cm}<{\centering}|m{0.7cm}<{\centering} | m{1.5cm}<{\centering} | m{1.5cm}<{\centering}  |m{1.5cm}<{\centering}|m{1.5cm}<{\centering} |
        m{1.5cm}<{\centering}|m{1.5cm}<{\centering} |
        m{1.5cm}<{\centering} | m{1.5cm}<{\centering} | m{1.5cm}<{\centering}}

            \hline\hline
            \multirow{2}{*}{\makecell{Method}}  & \multirow{2}{*}{\makecell{Year}} & \multicolumn{ 3}{|c}{Car} & \multicolumn{ 3}{|c}{Pedestrian} & \multicolumn{ 3}{|c}{Cyclist}   \\
            \cline{3-11}
              &  & Easy & Mod & Hard & Easy & Mod & Hard & Easy & Mod & Hard  \\
              \hline
              \multicolumn{ 11}{c}{Early-Fusion} \\
              \hline
              PFF3D \cite{wen2021fast} & 2021&	89.61&	85.08&	80.42&	48.74&	40.94&	38.54&	72.67&	55.71&	49.58\\
              \hline
              Painted PointRCNN \cite{vora2020pointpainting} &  2020 &	92.45&	88.11&	83.36&	58.70&	49.93&	46.29&	83.91&	71.54&	62.97 \\
              \hline
              PI-RCNN \cite{xie2020pi}&	2020&	91.44&	85.81&	81.00& - &- &- &- &- &- \\
              \hline
              Complexer-YOLO \cite{simon2019complexer}&	2019&	77.24&	68.96&	64.95&	21.42&	18.26&	17.06&	32.00&	25.43&	22.88\\
              \hline
              MVX-Net(PF) \cite{sindagi2019mvx} &	2019&	89.20&	85.90&	78.10& - &- &- &- &- &- \\
              \hline

              \multicolumn{ 11}{c}{Deep-Fusion} \\
              \hline
              RoIFusion \cite{chen2021roifusion}& 2021&	92.88&	89.03&	83.94&	46.21&	38.08&	35.97&	83.13&	67.71&	61.70\\
              \hline
              EPNet \cite{huang2020epnet}	&2020&	94.22&	88.47&	83.69&	- &- &- &- &- &- \\	
              \hline
              MAFF-Net \cite{zhang2020maff}&	2020&	90.79&	87.34&	77.66& 	- &- &- &- &- &- \\	
              \hline
              SemanticVoxels \cite{fei2020semanticvoxels}&	2020& - &- &- &	58.91&	49.93&	47.31&  - &- &- \\	
              \hline
              MVAF-Net \cite{wang2020multi}&	2020&	91.95&	87.73&	85.00&	- &- &- &- &- &- \\	
              \hline
              3D-CVF \cite{yoo20203d}&	2020&	93.52&	89.56&	82.45&	- &- &- &- &- &- \\	
              \hline
              MMF \cite{liang2019multi}&	2019&	93.67&	88.21&	81.99&	- &- &- &- &- &- \\
              \hline
              ContFuse \cite{liang2018deep}&	2018&	94.07&	85.35&	75.88& 	- &- &- &- &- &- \\
              \hline
              SparsePool \cite{wang2017fusing}& 2017&  - &- &- & 43.33&	34.15&	31.78&	43.55&	35.24&	30.15\\
              \hline

              \multicolumn{ 11}{c}{Late-Fusion} \\
              \hline
              CLOCs \cite{pang2020clocs} &	2020&	92.91&	89.48&	86.42&		- &- &- &- &- &- \\	
              \hline

              \multicolumn{ 11}{c}{Asymmetry-Fusion} \\
              \hline
              
              VMVS \cite{ku2019improving}& 2019& - &- &- & 60.34&	50.34&	46.45&	- &- &- \\	
              \hline
              MLOD \cite{deng2019mlod}&	2019&	90.25&	82.68&	77.97&	55.09&	45.40&	41.42&	73.03&	55.06&	48.21\\
              \hline
              MV3D \cite{chen2017multi}&	2017&	86.62&	78.93&	69.80&	- &- &- &- &- &- \\
              \hline

              \multicolumn{ 11}{c}{Weak-Fusion} \\
              \hline

              Faraway-Frustum \cite{zhang2020faraway}&	2020&	91.90&	88.08&	85.35&	52.15&	43.85&	41.68&	79.65&	64.54&	57.84\\
              \hline
              F-ConvNet \cite{wang2019frustum}&	2019&	91.51&	85.84&	76.11&	57.04&	48.96&	44.33&	84.16&	68.88&	60.05\\
              \hline
              IPOD \cite{yang2018ipod}&	2018&	86.93&	83.98&	77.85&	60.83&	51.24&	45.40&	77.10&	58.92&	51.01\\
              \hline
              F-PointNet \cite{qi2018frustum}&	2017&	91.17&	84.67&	74.77&	57.13&	49.57&	45.48&	77.26&	61.37& 53.78\\
              
            \hline\hline
        \end{tabular}
        }
    \end{center}
\end{table*}

\begin{table*}[h]
    \begin{center}
        \caption{Survey of 3D Task Results in KITTI\cite{Geiger2012CVPR} for Test Dataset}
        \label{tab:result-3d-kitti}
        \scalebox{0.75}{
        \begin{tabular}
        {m{4cm}<{\centering}|m{0.7cm}<{\centering} | m{1.5cm}<{\centering} | m{1.5cm}<{\centering}  |m{1.5cm}<{\centering}|m{1.5cm}<{\centering} |
        m{1.5cm}<{\centering}|m{1.5cm}<{\centering} |
        m{1.5cm}<{\centering} | m{1.5cm}<{\centering} | m{1.5cm}<{\centering}}

            \hline\hline
            \multirow{2}{*}{\makecell{Method}}  & \multirow{2}{*}{\makecell{Year}} & \multicolumn{ 3}{|c}{Car} & \multicolumn{ 3}{|c}{Pedestrian} & \multicolumn{ 3}{|c}{Cyclist}   \\
            \cline{3-11}
              &  & Easy & Mod & Hard & Easy & Mod & Hard & Easy & Mod & Hard  \\
              \hline
              \multicolumn{ 11}{c}{Early-Fusion} \\
              \hline
              PFF3D \cite{wen2021fast}& 2021& 81.11&	72.93&	67.24	&43.93	&36.07&	32.86&	63.27	&46.78&	41.37\\
              \hline
              Painted PointRCNN \cite{vora2020pointpainting}&  2020 & 82.11 &	71.70 &	67.08 & 50.32 & 40.97	&	37.87 &	77.63 &	63.78 &	55.89 \\
              \hline
              PI-RCNN \cite{xie2020pi} &	2020&	84.37&	74.82&	70.03& - &- &- &- &- &- \\
              \hline
              Complexer-YOLO \cite{simon2019complexer}&	2019& 55.93	&47.34&	42.60&	17.60&	13.96&	12.70&	24.27&	18.53&	17.31\\
              \hline
              MVX-Net(PF) \cite{sindagi2019mvx} &	2019&	83.20&	72.70&	65.20& - &- &- &- &- &- \\
              \hline

              \multicolumn{ 11}{c}{Deep-Fusion} \\
              \hline
              RoIFusion \cite{chen2021roifusion}& 2021& 88.09&	79.36&	72.51& 42.22&	35.14&	32.92&	80.84&	64.05&	58.37 \\
              \hline
              EPNet \cite{huang2020epnet}	&2020&	89.81&	79.28&	74.59&	- &- &- &- &- &- \\	
              \hline
              MAFF-Net \cite{zhang2020maff}&	2020&	85.52&	75.04&	67.61& 	- &- &- &- &- &- \\	
              \hline
              SemanticVoxels \cite{fei2020semanticvoxels}&	2020& - &- &- &	50.90&	42.19&	39.52& - &- &- \\	
              \hline
              MVAF-Net \cite{wang2020multi}&	2020&	87.87&	78.71&	75.48&	- &- &- &- &- &- \\	
              \hline
              3D-CVF \cite{yoo20203d}&	2020&	89.20&	80.05&	73.11&	- &- &- &- &- &- \\	
              \hline
              MMF \cite{liang2019multi}&	2019&	88.40&	77.43&	70.22&	- &- &- &- &- &- \\
              \hline
              SCANet \cite{lu2019scanet}&	2019&	76.09&	66.30&	58.68& - &- &- &- &- &- \\
              \hline
              ContFuse \cite{liang2018deep}&	2018&	83.68&	68.78&	61.67& 	- &- &- &- &- &- \\
              \hline
              PointFusion \cite{xu2018pointfusion}&	2018&	77.92&	63.00&	53.27& 	- &- &- &- &- &- \\
              \hline
              SparsePool \cite{wang2017fusing}& 2018&  - &- &- & 37.84&	30.38&	26.94&	40.87&	32.61&	29.05\\
              \hline

              \multicolumn{ 11}{c}{Late-Fusion} \\
              \hline
              CLOCs \cite{pang2020clocs} &	2020&	89.16&	82.28&	77.23&		- &- &- &- &- &- \\
              \hline

              \multicolumn{ 11}{c}{Asymmetry-Fusion} \\
              \hline
              
              VMVS \cite{ku2019improving}& 2019& - &- &- & 53.44&	43.27&	39.51&	- &- &- \\	
              \hline
              MLOD \cite{deng2019mlod}&	2019 &	77.24&	67.76&	62.05&	47.58&	37.47&	35.07&	68.81&	49.43&	42.84\\
              \hline
              MV3D \cite{chen2017multi}&	2017&	74.97&	63.63&	54.00&	- &- &- &- &- &- \\
              \hline
              \multicolumn{ 11}{c}{Weak-Fusion} \\
              \hline

              Faraway-Frustum \cite{zhang2020faraway}&	2020&	87.45&	79.05&	76.14&	46.33&	38.58&	35.71&	77.36&	62.00&	55.40\\
              \hline
              F-ConvNet \cite{wang2019frustum}&	2019& 87.36&	76.39&	66.69 & 52.16&	43.38&	38.80&	81.98& 65.07&	56.54\\
              \hline
              IPOD \cite{yang2018ipod}&	2018& 79.75&	72.57&	66.33&	56.92&	44.68&	42.39&	71.40&	53.46&	48.34\\
              \hline
              F-PointNet \cite{qi2018frustum}&	2018&	82.19&	69.79&	60.59&	50.53&	42.15&	38.08&	72.27&	56.12&	49.01\\

            \hline\hline
        \end{tabular}
        }
    \end{center}
\end{table*}

\subsection{Strong-fusion}

\begin{figure}[htbp]
\centering
\includegraphics[width=0.48\textwidth]{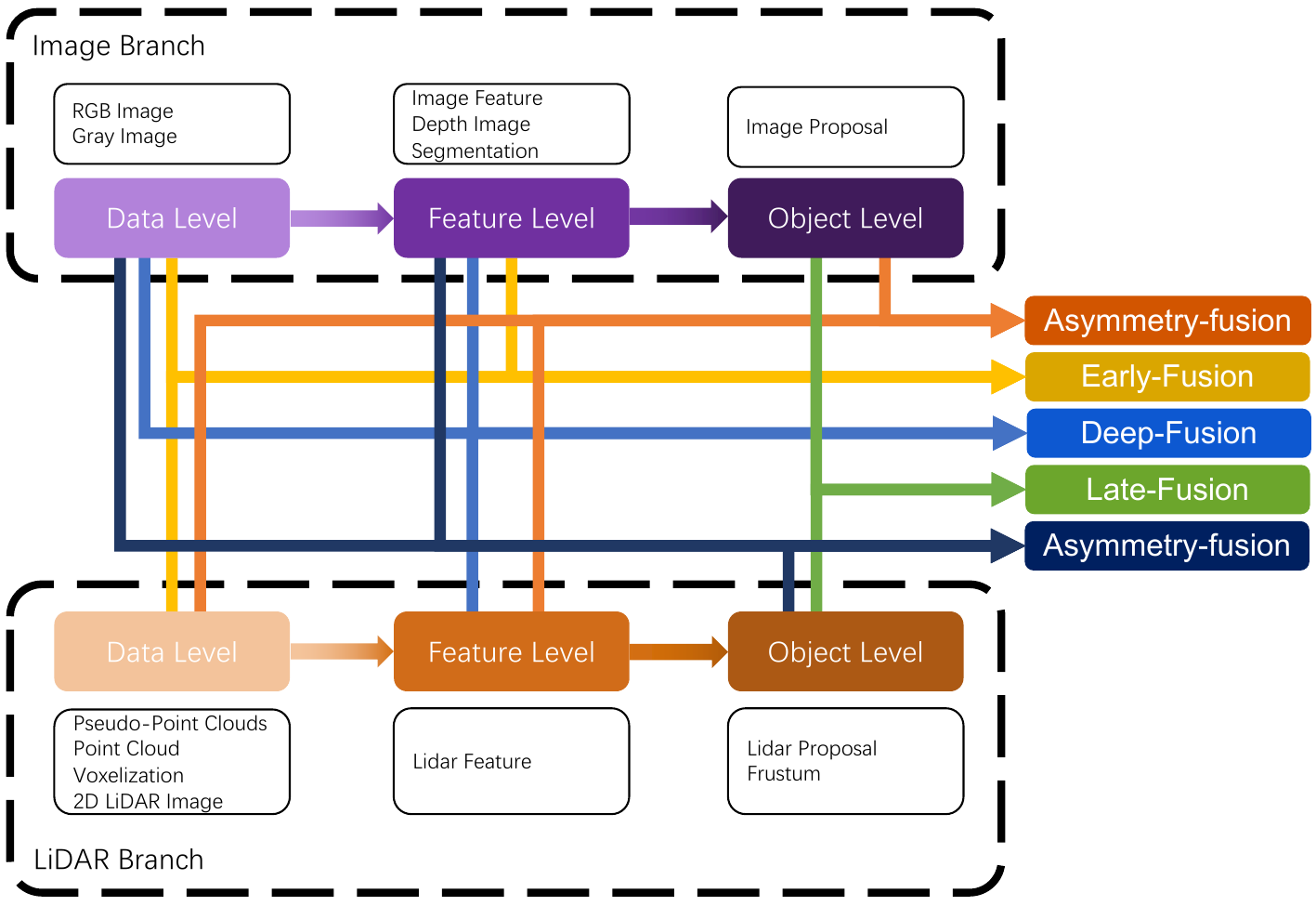}
\caption{Strong-Fusion Overview}
\label{fig:strong-fusion}
\end{figure}

We divide the strong-fusion into four categories, as early-fusion, deep-fusion, late-fusion, and asymmetry-fusion, by the different combination stages of LiDAR and camera data representations. As the most studied fusion method, strong fusion shows a lot of outstanding achievements in recent years \cite{pang2020clocs,vora2020pointpainting,wang2021pointaugmenting}. From the overview in Figure  \ref{fig:strong-fusion}, it is easy to notice that each minor class in strong-fusion is highly dependent on the LiDAR point cloud instead of the camera data. We will then discuss each of them in particular.

\begin{figure}[htbp]
\centering
\includegraphics[width=0.40\textwidth]{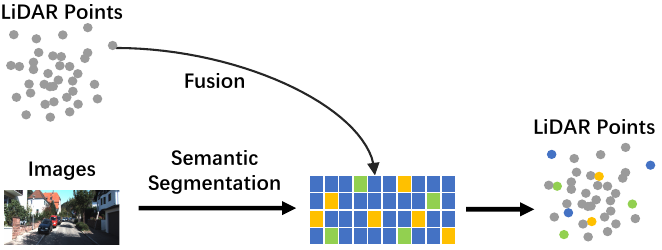}
\caption{An Example of Early-Fusion}
\label{fig:early-fusion}
\end{figure}

\textbf{Early-fusion.} Unlike the traditional definition of data-level-fusion, which is a method that fuses data in each modality directly by a spatial alignment and projection at the raw data level, early-fusion fuses LiDAR data at the data level and camera data at data-level or feature-level. One example of early-fusion can be the model in Figure \ref{fig:early-fusion}.

As for the LiDAR branch mentioned above, the point cloud can be used in the form of 3D points with reflectance, voxelized tensor, front-view/ range-view/ bird's eye view, as well as pseudo-point clouds. Though all these data have different intrinsic characteristics, which are highly associated with the latter LiDAR backbone, most of these data are produced with a rule-based procession except pseudo-point clouds\cite{wang2020kda3d}. Besides, all these data representations of LiDAR can be visualized straightforward because the data in this stage still have interpretability compared to the embedding in feature space.

As for the image path, the strict data-level definition should only contain data as RGB or Gray, which lacks generality and rationality. Compared to the traditional definition of early-fusion, we here loosen camera data to data-level and feature-level data. Especially, here we treat semantic segmentation task results in image branch benefiting the 3D object detection as feature-level representation since these kinds of "object-level" features are different from the final object-level proposals for the whole task.

\cite{vora2020pointpainting} and \cite{xie2020pi} fused semantic-feature in image branch and raw lidar point clouds together which results in a better performance in object detection tasks. \cite{simon2019complexer}, and \cite{dou2019seg} also utilize semantic-feature, but different from the above methods, it pre-processes raw lidar point cloud into voxelized tensor to further leverage a more advanced LiDAR backbone. \cite{meyer2019sensor} transforms 3D lidar point clouds into the 2D image and fuses feature-level representation in the image branch leveraging mature CNN techniques to achieve better performance. \cite{wen2021fast} fused raw RGB pixel with the voxelized tensor while \cite{wang2020kda3d} directly combines pseudo-point clouds generated from the image branch and raw point clouds from the LiDAR branch together to accomplish object detection tasks.

Based on the VoxelNet\cite{zhou2018voxelnet}, \cite{sindagi2019mvx} proposes one of the fusion methods as point-fusion, which directly attached the image feature vector of the corresponding pixel to the voxelized vector. \cite{xu2018pointfusion} proposes the dense-fusion that attaches every original point with the global features from the image branch. \cite{melotti2018multimodal} focuses on 2D pedestrian detection using CNN. As early-fusion, it directly fuses different branches before inputting into CNN. \cite{zhang2020maff} proposes one fusion method named point attention fusion which fused image feature to a voxelized tensor in LiDAR point clouds.

\begin{figure}[htbp]
\centering
\includegraphics[width=0.38\textwidth]{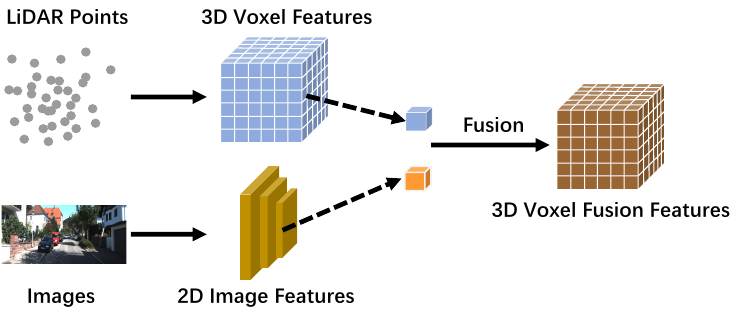}
\caption{An Example of Deep-Fusion}
\label{fig:deep-fusion}
\end{figure}

\textbf{Deep-fusion.} Deep-fusion methods fuse cross-modal data at the feature level for the LiDAR branch but data-level and feature-level for the image branch. For example, some methods use feature extractor to acquire the embedding representation of LiDAR point cloud and camera image respectively and fuse the feature in two modalities by a series of downstream modules \cite{yoo20203d, huang2020epnet}. However, unlike other strong-fusion methods, deep-fusion sometimes fuses features in a cascading way \cite{huang2020epnet,bijelic2020seeing, liang2018deep}, which both leverage raw and high-level semantic information. One example of deep-fusion can be the model in Figure \ref{fig:deep-fusion}.

\cite{xu2018pointfusion} proposes the global-fusion attached global LiDAR feature with the global features from the image branch. \cite{sindagi2019mvx} proposes the other fusion methods as voxel-fusion, which attached the ROI pooling image feature vector to a dense feature vector for each voxel in LiDAR point clouds.  \cite{zhang2020maff} proposes another method named dense attention fusion that fuses pseudo images from multiple branches. \cite{lu2019scanet,liang2019multi}, propose two deep-fusion method each. EPNet \cite{huang2020epnet} a deep LIDAR-Image fusion estimating the importance of corresponding image features to reduce the noisy influence. \cite{bijelic2020seeing} presents a multi-modal data set in extreme weather and fused each branch in a deep-fusion way, which greatly improves the robustness of the autonomous driving model. Other deep-fusion work includes \cite{yoo20203d,liang2018deep,daniel2017fast,chen2021roifusion,wang2020multi,wang2017fusing,kim2018robust,cheng2019noise,theodose2020r,fei2020semanticvoxels} which have the seemingly same fusion module.

\begin{figure}[htbp]
\centering
\includegraphics[width=0.40\textwidth]{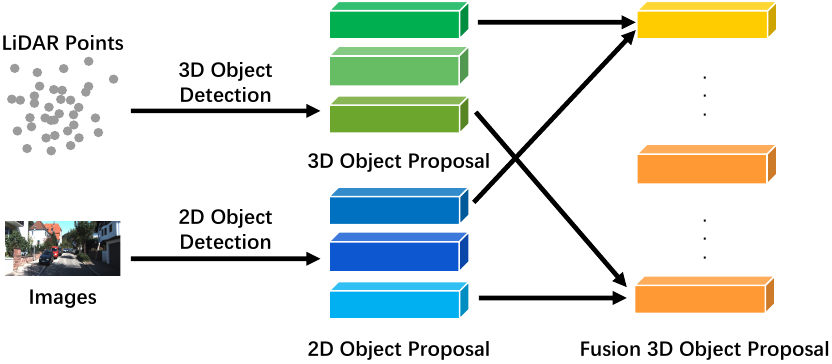}
\caption{An Example of Late-Fusion}
\label{fig:late-fusion}
\end{figure}

\textbf{Late-fusion.} Late-fusion, also known as object-level fusion, denotes the methods that fuse the result of pipelines in each modality. For example, some late-fusion methods leverage output from both the LiDAR point cloud branch and camera image branch and make the final prediction based on the result in two modalities\cite{pang2020clocs}. Notice that both branch proposals should have the same data format as the final results but vary in quality, quantity, and precision. Late-fusion can be regarded as a kind of ensemble method that utilizes multi-modal information to optimize the final proposal. One example of late-fusion can be the model in Figure \ref{fig:late-fusion}.

As mentioned above, \cite{pang2020clocs} leverages late-fusion for secondly refining every 3D region proposal's score combining 2D proposal in image branch with the 3D proposal in LiDAR branch. Besides, for every overlapping region, it utilized statistical features like confidence score, distance and IoU. \cite{asvadi2018multimodal} focuses on 2D object detection, it combines the proposals from two branches along with features like confidence score, and the model outputs the final IoU score. \cite{gu2019road},\cite{gu2019integrating} solves the road detection by combing the segmentation results together. As late-fusion in \cite{melotti2018multimodal}, it summarizes the scores from different branches for the same 3D detection proposal into one final score. 

\begin{figure}[htbp]
\centering
\includegraphics[width=0.38\textwidth]{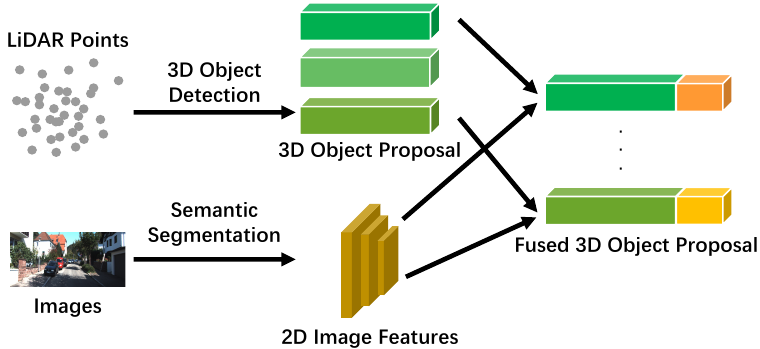}
\caption{An Example of Asymmetry-Fusion}
\label{assy}
\end{figure}

\textbf{Asymmetry-fusion.} Besides the early-fusion, deep-fusion, and late-fusion, some methods treat cross-modal branches with different privileges, so we define methods that fuse object-level information from one branch while data-level or feature-level from other branches as asymmetry-fusion. Unlike other methods in strong-fusion, which treat both branches in seemingly equal status, asymmetry-fusion has at least one branch dominating while other branches provide auxiliary information to conduct the final task. One example of late-fusion can be the model in Figure \ref{assy}. Especially compared to late-fusion, though they might have the same extract feature using proposal \cite{pang2020clocs}, asymmetry-fusion only have one proposal from one branch while late-fusion have proposals from all the branches.

Such fusing methods are reasonable due to the excellent performance using convolutional neural network on camera data which filters semantically useless points in the point cloud and promotes the performance on 3D LiDAR backbone in a frustum perspective, such as \cite{zhao20193d}. It extracts frustum in raw point clouds along with corresponding pixel's RGB information to output the parameters of the 3D bounding box. However, some works think outside the box and use the LiDAR backbone to guide the 2D backbone in a multi-view style and achieve higher accuracy. \cite{ku2019improving} focuses on pedestrian detection by the extracted multi-view image based on 3D detection proposals, which further utilizes  CNN to refine the previous proposal. \cite{chen2017multi} and \cite{deng2019mlod} refines the 3D proposal predicted solely by the LiDAR branch with ROI feature in other branches. \cite{braun2016pose} focuses on 2D detection, utilizing 3D region proposal from LiDAR branch and re-projecting to 2D proposal along with the image features for further refinement. \cite{chen20173d} proposes a 3D potential bounding box by statistical and rule-based information. Combining with the image feature, it outputs the final 3D proposal.\cite{singh2020lidar} is focusing on small object detection accomplished by a specially collected data set, which is essentially a 2D semantic segmentation task combining the proposal from LiDAR with the raw RGB image to output the final results.

\subsection{Weak-fusion}

\begin{figure}[htbp]
\centering
\includegraphics[width=0.35\textwidth]{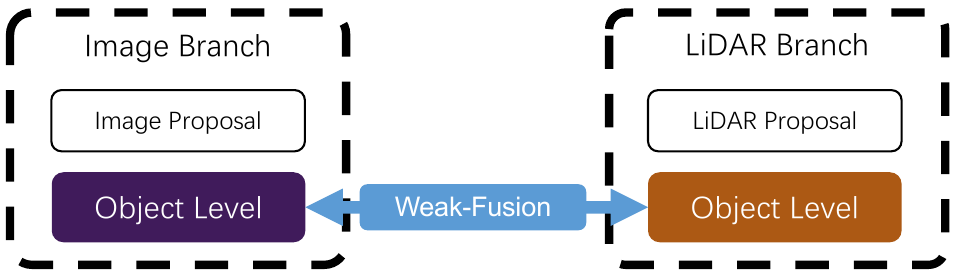}
\caption{Weak-Fusion Overview}
\label{fig:weak-fusion}
\end{figure}

Unlike strong-fusion, weak-fusion methods do not fuse data/ feature/ object directly from branches in multi-modalities but operate data in other ways. The weak-fusion-based methods commonly use rule-based methods to utilize data in one modality as a supervision signal to guide the interaction of another modality. Figure \ref{fig:weak-fusion} demonstrates the basic framework of the weak-fusion schema. For example, the 2D proposal from the CNN in the image branch might result in a frustum in the raw lidar point cloud. However, unlike combining image features as asymmetry-fusion mentioned above, weak fusion directly  input those raw LiDAR point cloud selected into the LiDAR backbone to output the final proposal \cite{qi2018frustum}.

\cite{wang2019frustum} advance the techniques  by dividing each frustum into several parts by fixed chosen stride which, further increases the 3D detection accuracy. \cite{zhang2020faraway} focuses on remote sparse-point-cloud object detection. \cite{yang2018ipod} filters out all the background points of LiDAR point cloud in frustum from semantic segmentation results in an image. \cite{tang2019transferable} focuses on semi-supervised and transfer learning where frustum is proposed according to 2D image proposal.

Other weak-fusion like \cite{dorka2020modality} highlights the real-time detection performance of 2D objects by selecting only one model of the two branches at each time to predict the final proposal using a reinforcement learning strategy. In \cite{du2018general}, multiple  3D box proposals are generated by 2D detection proposals in the image branch, and then the model outputs the final 3D detection box with its detection score. \cite{shin2019roarnet} uses an image to predict 2D bounding box and 3D pose parameters and further refines it utilizing the LiDAR point clouds in the corresponding area.

\subsection{Other Fusion Methods}

Some work could not be simply defined as any kind of fusion mentioned above because they possess more than one fusion method in the whole model framework, such as a combination of deep-fusion and late-fusion \cite{ku2018joint} while \cite{wang2021pointaugmenting} combines early-fusion and deep-fusion together. These methods suffer from redundancy at the model design view, which is not the mainstream for fusion modules.

\section{Opportunities in Multi-Modal Fusion}

Multi-modal fusion methods for perception tasks in autonomous driving have achieved rapid progress in recent years, varying from more advanced feature representations to more complex deep learning models\cite{wang2021multi, cui2021deep}. However, there remains some more open problems to be fixed. We here summarize some critical and essential work to be done in the future into the following aspects.

\subsection{More Advanced Fusion Methodology}

Current fusion models suffer from the problem of misalignment and information loss \cite{shin2019roarnet,yang2019automatic,chen2021cross}. Besides, the flat fusion operations \cite{vora2020pointpainting,dou2019seg} also prevent the further improvement of perception task performance. We summarize them as two aspects: Misalignment and Information Loss, More Reasonable Fusion Operations.

\textbf{Misalignment and Information Loss}

The intrinsic and extrinsic of the camera and LiDAR are vastly different from each other. Data in both modalities need to be re-organized under a new coordinate system. Traditional early and deep fusion methods utilize an extrinsic calibration matrix to project all the LiDAR points directly to the corresponding pixels or vice versa \cite{meyer2019sensor,vora2020pointpainting,sindagi2019mvx}. However, this point-by-pixel alignment is not accurate enough because of the sensory noise. Therefore, we can see that in addition to such strict correspondence, some work \cite{xie2020pi} utilizing the surrounding information as a supplement results in better performance. 

Besides, there exists some other information loss during the transformation of input and feature space. Generally, the projection of dimension reduction operation will inevitably lead to massive information loss, e.g., mapping the 3D LiDAR point cloud into the 2D BEV image. Therefore, by mapping the two modal data into another high-dimensional representation specially designed for fusion, future works can efficiently utilize the raw data with less information loss.

\textbf{More Reasonable Fusion Operations}

The current research works use intuitive methods to fuse cross-modal data, such as concatenation and element-wise multiply \cite{sindagi2019mvx,wang2021pointaugmenting}. These simple operations may not be able to fuse the data with a large distribution discrepancy, and as a result, hard to close the semantic gap between two modalities. Some work tries to use a more elaborated cascading structure to fuse the data and improve performance \cite{chen2017multi,liang2018deep}. In future research, mechanisms such as Bi-linear mapping\cite{fukui2016multimodal,kim2016hadamard,ben2017mutan} can fuse the feature with different characteristics.

\subsection{Multi-Source Information Leverage}

Single frame at the front view is the typical scenario for autonomous driving perception tasks\cite{Geiger2012CVPR}. However, most frameworks utilize limited information without elaborated designed auxiliary tasks to further understand the driving scenes. We summarize them as With More Potential Useful Information and Self-Supervision for Representation Learning.

\textbf{With More Potential Useful Information}

Existing methods \cite{wang2021multi} lack the effective use of the information from multiple dimensions and sources. Most of them focus on a single frame of multi-modal data at the front view. As a result, other meaningful information is under-utilized such as semantic, spatial, and scene contextual information.

Some models \cite{vora2020pointpainting,xie2020pi,dou2019seg} try to use the results obtained from the image semantic segmentation task as additional features, while others may exploit the features in intermediate layers of a neural network backbone whether trained by specific downstream tasks or not \cite{liang2018deep}. In autonomous driving scenarios, many downstream tasks with explicit semantic information may greatly benefit the performance of object detection tasks. For example, lane detection can intuitively provide additional help for detecting vehicles between the lanes, and the semantic segmentation results can improve object detection performance \cite{vora2020pointpainting,xie2020pi,dou2019seg}. Therefore, future research can jointly build a complete semantic understanding framework of the cityscape scenarios through various downstream tasks like detecting the lane, traffic light, and signs to assist perception tasks' performance.

In addition, current perception tasks mainly rely on a single frame that overlooks the temporal information. Recent LiDAR-based method \cite{qi2021offboard} combines a series of frames to improve the performance. The time-series information contains serialized supervision signals, which can provide more robust results than methods that use a single frame.

Therefore, future work may dig deeper into utilizing temporal, contextual, and spatial information for continuous frames with innovative model designs.

\textbf{Self-Supervision for Representation Learning}

The mutual-supervised signals naturally exist among the cross-modal data sampled from the same real-world scenarios but different perspectives. However, current methods cannot mine the co-relationship among each modality, suffering from lacking a deep understanding of data. In the future, the studies can be focused on how to use the multi-modal data for self-supervised learning, including pre-training, fine-tuning or contrastive learning. By implementing these state-of-the-art mechanisms, the fusion model will result in a deeper understanding of the data and achieve better results, which has already shown some promising signs in other areas while leaving a blank space for autonomous driving perception \cite{liu2020p4contrast}.

\subsection{Intrinsic Problems in Perception Sensors}

Domain bias and resolution are highly related to the real-world scenes and the sensors \cite{Geiger2012CVPR}. These unexpected flaws prevent large-scale training and implementation for autonomous driving deep learning models, which need to be solved in future work. 

\textbf{Data Domain Bias}

In autonomous driving perception scenes, the raw data extracted by different sensors is accompanied by severe domain-related characteristics. Different camera systems have their optical properties, while LiDAR may vary from mechanical LiDAR to solid-state LiDAR. What is more, data itself may be domain-biased such as weather, season, or location \cite{sun2020scalability, caesar2020nuscenes}, even if it is captured by the same sensors. As a result, the detection model can not be adapted to new scenarios smoothly. Such defects prevent the collection of a large-scale dataset and the re-usability of the original training data due to the failures of generalization. Therefore, it is crucial to find a way to eliminate domain bias and integrate different data sources adaptively in future work.

\textbf{Conflicts with Data Resolution}

Sensors from different modalities often have different resolutions \cite{langer2020domain, yi2021complete}. For example, the spatial density of LiDAR is notably lower than that of the image. No matter what projection method is adopted, some information is eliminated because the corresponding relationship cannot be found. This may lead the model to be dominated by the data of one specific modality, whether it is due to the different resolution of the feature vector or the imbalance of original information. Therefore, future works can explore a new data representation system compatible with sensors in different spatial resolutions.

\section{Conclusion}

In this paper, we review over 50 related papers about the multi-modal sensor fusion for autonomous driving perception tasks. To be specific, we first propose an innovative way to classify these papers into three classes by a more reasonable taxonomy from the fusion perspective. Then we conduct an in-depth survey about the data format and representation of the LiDAR and camera and describe the different characteristics. Finally, we make a detailed analysis about the remaining problems of multi-modal sensor fusion and introduce several new possible directions, which may enlighten future research works.

{\small
\bibliographystyle{ieee_fullname}
\bibliography{cvpr}
}

\end{document}